%% file: main.tex
\documentclass[journal]{vgtc}                     

\title{EgoTrigger: Toward Audio-Driven Image Capture for Human Memory Enhancement in All-Day Energy-Efficient Smart Glasses}

\author{%
  \authororcid{
  Akshay Paruchuri}{0000-0003-4664-3186},
  \authororcid{
  Sinan Hersek}{0000-0001-7333-005X},
  \authororcid{
  Lavisha Aggarwal}{0009-0006-2280-1302},
  \authororcid{
  Qiao Yang}{0009-0008-9304-3654},
  \authororcid{
  Xin Liu}{0000-0002-9279-5386},\\
  \authororcid{
  Achin Kulshrestha}{0009-0004-8632-9575},
  \authororcid{
  Andrea Colaco}{0009-0001-6661-2216},
  \authororcid{
  Henry Fuchs}{0000-0002-8834-4638}, and 
  \authororcid{
  Ishan Chatterjee}{0000-0002-2123-6392}
}

\authorfooter{
  \item
  	Akshay Paruchuri and Henry Fuchs are with UNC Chapel Hill.\\
  	Contact: \{akshay, fuchs\}@cs.unc.edu
  \item
  	Ishan Chatterjee and the remaining co-authors are with Google. \\
  	Contact: ishanc@google.com
}

\input{sections/main/0-abstract}


\teaser{
  \centering
  \includegraphics[width=\linewidth]{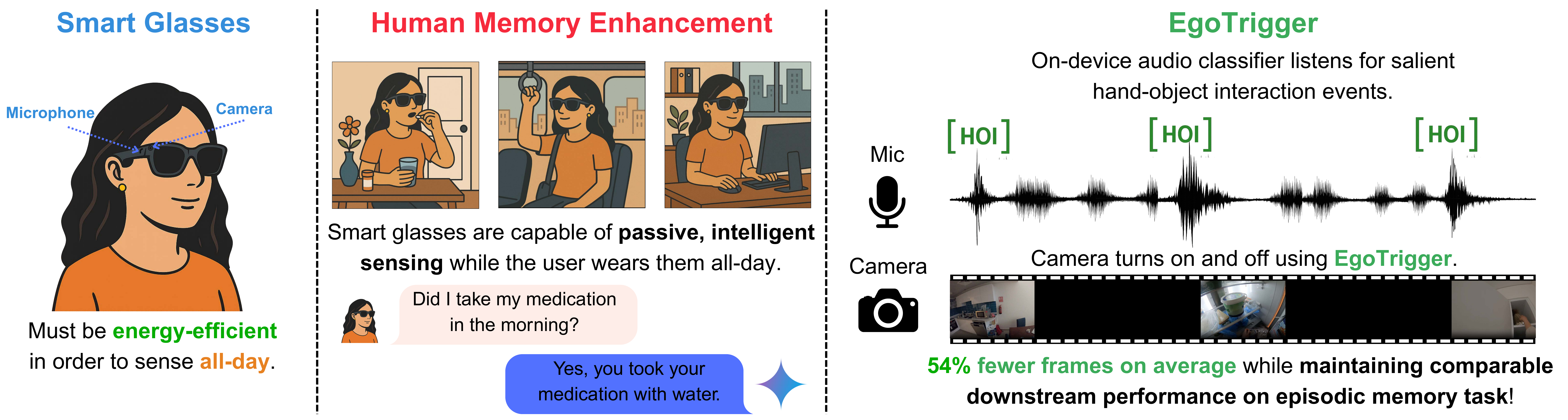}
  \caption{We present \emph{EgoTrigger}, an image capture approach designed for resource-constrained smart eyeglass systems. \emph{EgoTrigger} uses a lightweight audio classification model and a custom classification head to trigger image capture from hand-object interaction (HOI) audio cues, such as the sound of a medication bottle being opened. Subsequently, image capture can be triggered off either based on a fixed duration (e.g., 1 second) or a hysteresis-based approach. \emph{EgoTrigger} is capable of using 54\% fewer frames on average while maintaining or exceeding performance on an episodic memory task, signifying potential for energy-efficient usage as a part of all-day smart glasses capable of human memory enhancement.}
  \label{fig:teaser}
}

\graphicspath{{figs/}{figures/}{pictures/}{images/}{./}} 

\usepackage{tabu}                      
\usepackage{booktabs}                  
\usepackage{lipsum}                    
\usepackage{mwe}                       

\usepackage{amsmath}

\usepackage[dvipsnames,svgnames]{xcolor}
\usepackage{tcolorbox}

\usepackage{graphicx}   

\usepackage{mathptmx}                  

\usepackage{multirow}    
\usepackage{tabularx}
\tcbuselibrary{skins,breakable} 

\begin{document}

\input{sections/main/1-introduction}
\input{sections/main/2-related_work}
\input{sections/main/3-our_approach}
\input{sections/main/4-HME-QA}
\input{sections/main/5-experimental_setup}
\input{sections/main/6-evaluation}
\input{sections/main/7-discussion}
\input{sections/main/8-limitations_and_future_work}
\input{sections/main/9-conclusion}
\input{sections/main/10-acknowledgments}

\clearpage

\bibliographystyle{abbrv-doi}

\bibliography{main}

\clearpage

\input{sections/supp/0-all_supp}

\end{document}

%% file: sections/main/0-abstract.tex
\abstract{%
    All-day smart glasses are likely to emerge as platforms capable of continuous contextual sensing, uniquely positioning them for unprecedented assistance in our daily lives. Integrating the multi-modal AI agents required for human memory enhancement while performing continuous sensing, however, presents a major energy efficiency challenge for all-day usage. Achieving this balance requires intelligent, context-aware sensor management. Our approach, \emph{EgoTrigger}, leverages audio cues from the microphone to selectively activate power-intensive cameras, enabling efficient sensing while preserving substantial utility for human memory enhancement. \emph{EgoTrigger} uses a lightweight audio model (YAMNet) and a custom classification head to trigger image capture from hand-object interaction (HOI) audio cues, such as the sound of a drawer opening or a medication bottle being opened. In addition to evaluating on the QA-Ego4D dataset, we introduce and evaluate on the Human Memory Enhancement Question-Answer (HME-QA) dataset. Our dataset contains 340 human-annotated first-person QA pairs from full-length Ego4D videos that were curated to ensure that they contained audio, focusing on HOI moments critical for contextual understanding and memory. Our results show \emph{EgoTrigger} can use 54\% fewer frames on average, significantly saving energy in both power-hungry sensing components (e.g., cameras) and downstream operations (e.g., wireless transmission), while achieving comparable performance on datasets for an episodic memory task. We believe this context-aware triggering strategy represents a promising direction for enabling energy-efficient, functional smart glasses capable of all-day use --- supporting applications like helping users recall where they placed their keys or information about their routine activities (e.g., taking medications).
}

%% file: sections/main/1-introduction.tex
\firstsection{Introduction}

\maketitle

As an all-day wearable, smart glasses provide a unique platform to enable continuous contextual sensing from an egocentric perspective. This long-context sensing can be processed by powerful agentic multi-modal foundation models (MLLMs) to enable unprecedented assistance in our daily lives. For example, researchers have long envisioned enhancing users' memory of their daily actions via wearables and glasses~\cite{rhodes1997, hoisko2000, devaul2003}. Users may eventually be able to ask in natural language, "Where did I leave my keys earlier?" or "Did I turn the oven off before leaving?", among other things.

However, realizing this vision of memory enhancement hinges on overcoming significant energy efficiency hurdles. As face-worn devices, smart glasses face strict form-factor limitations, constraining battery size, thermal dissipation, and onboard computation, and making all-day operation challenging \cite{billinghurst1999}. Even the most optimized, highly capable MLLMs (e.g., Gemini Nano~\cite{team2023gemini}) contain over a billion parameters, making local deployment intractable on smart glasses hardware. Rather, these MLLMs run on powerful processors in the cloud~\cite{team2023gemini, achiam2023gpt, anthropic2024claude} with the glasses passing their sensor data over wireless link by way of a connected phone (e.g. Vuzix M400~\cite{vuzix_m400_smart_glasses}, Ray-Ban Meta~\cite{RaybanMeta}) or separate compute puck (e.g., Meta's Orion Prototype~\cite{meta_orion_2024}).

Despite offloading inference, a critical energy bottleneck for memory enhancement remains: continuous environmental sensing and wireless transmission to the agent, particularly from power-hungry and high-bandwidth cameras required for visual understanding \cite{khan2016sensors}. To enable passive, intelligent sensing for applications like memory enhancement without constant battery drain, smarter sensing strategies are required.

We hypothesize that inexpensive sensing modalities (e.g., audio from a microphone) will be readily available in future smart glasses and can be leveraged to effectively trigger capture of imagery from the camera. Prior research underscores the significance of hand-object interactions (HOIs) in shaping our memories, emphasizing that bodily engagement, particularly involving our hands, plays a crucial role in how we encode, recall, and structure key life experiences~\cite{iani2019embodied}. Audio cues, such as those indicating an HOI, can therefore serve as reliable indicators for moments when capturing visual information is most pertinent.

Building on this insight, we introduce \textbf{\emph{EgoTrigger}}, an energy-efficient approach for episodic memory enhancement using audio-gated visual capture. \emph{EgoTrigger} employs a lightweight audio classification model to detect salient audio events (e.g., HOI sounds like opening a medication bottle) and trigger brief camera captures. This selective capture strategy significantly reduces the need for continuous camera operation, maintaining performance on challenging human memory enhancement baselines. Our evaluations show \emph{EgoTrigger} is capable of using 54\% fewer frames on average while achieving comparable performance on downstream episodic memory question-answering tasks compared to continuous capture and fixed decimation baselines.

The primary contributions of this work are:
\begin{itemize}
    \item We present \emph{EgoTrigger}, a novel framework that uses audio cues from hand-object interactions (HOIs) to intelligently gate power-intensive cameras, such as those that would be a part of smart glasses. We validate its practical viability through an analysis of HOI audio event detection, its robustness to environmental noise, and a real-world assessment of its false positive rate on continuous video streams.
    \item We demonstrate that this audio-driven strategy reduces required visual frames by 54\% on average while maintaining downstream episodic memory task performance within 2\% of continuous capture, achieving a near Pareto-optimal balance between utility and energy savings.
    \item We introduce and release the Human Memory Enhancement Question-Answer (HME-QA) dataset, the first multimodal egocentric QA dataset curated for HOI-centric scenarios with guaranteed, paired audio, to facilitate future research.
\end{itemize}

Our code and access to the HME-QA dataset annotations can be found on our project page: \url{https://egotrigger.github.io/}

\begin{figure*}[t!]
    \centering
    \includegraphics[width=1\textwidth]{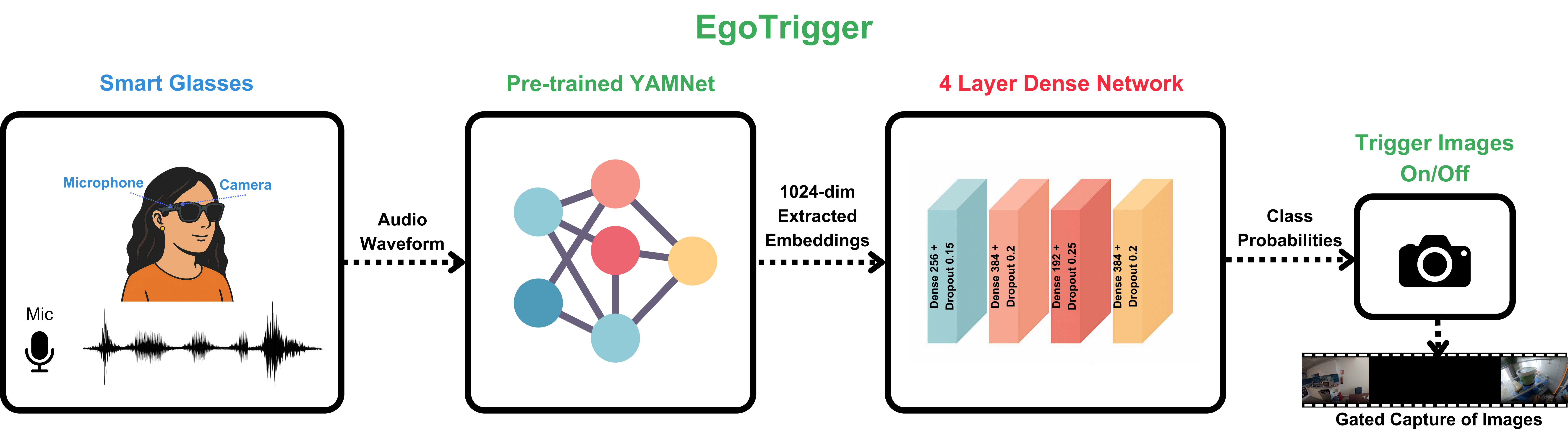}
    \vspace{-2em}
    \caption{\textbf{EgoTrigger.} Our approach focuses on audio from smart glasses that are processed by a pre-trained YAMNet model to extract 1024-dimensional embeddings, which are passed through a 4-layer dense network that serves as our custom, binary classification head. The resulting class probabilities for $C_0$ (indicates no hand-object interaction, or HOI) and $C_1$ (HOI) are used to trigger image capture, enabling gated visual sensing based on salient HOI audio events.}
    \label{fig:egotrigger_arch}
    \vspace{-0.4cm}
\end{figure*}

%% file: sections/main/2-related_work.tex
\section{Related work}

\noindent\textbf{Energy-Efficient Sensing for Wearables.} Wearable, network connected devices have become increasingly prevalent in our daily lives and typically feature a battery that needs to be recharged~\cite{qaim2020towards}. Such devices, which most commonly include smart watches today, are expected to last all day akin to our smartphones~\cite{khan2016sensors} in order to be usable. Prior works on improving energy-efficiency in wearables include device-specific optimization~\cite{kalantarian2015power}, energy harvesting~\cite{bahk2015flexible}, and adaptive device management frameworks~\cite{qi2013adasense, kindt2015adaptive, sunder2025smartapm}. Frameworks like AdaSense~\cite{qi2013adasense} and SmartAPM~\cite{kindt2015adaptive} have demonstrated significant energy savings in wearables by dynamically adjusting sensor sampling rates and leveraging deep reinforcement learning for power management, achieving up to 51\% and 36\% energy reductions respectively while preserving user experience. Approaches specific to device capabilities such as imaging have also been explored - for example, event cameras offer compelling advantages such as low latency, high dynamic range, and reduced power consumption, but are currently hindered by challenges including high costs, limited availability, and the need for specialized algorithms~\cite{chakravarthi2024recent}.

Smart glasses in particular face significant energy constraints due to their compact form factors, limited battery capacities, and the continuous operation of power-intensive sensors like cameras~\cite{billinghurst1999}. These constraints continue to be limiting for present day devices, such as the Vuzix M400~\cite{vuzix_m400_smart_glasses} and Ray-Ban Meta~\cite{RaybanMeta}, typically requiring some kind of "split compute" system across a separate phone or puck in order to enable meaningful user experiences with resource-constrained smart glasses. Researchers have optimized energy use in XR devices from the perspective that such devices will feature power-hungry displays that subsequently can be optimized using perceptually guided techniques such as adaptive shading, adaptive frame rates, luminance clipping, and color foveation~\cite{jindal2021perceptual, chen2024pea}. A co-optimization framework that balances energy consumption, latency, and accuracy in specific tasks such as scene reconstruction~\cite{tian2024towards} has also been proposed and shows promise for holistic system design in energy-constrained XR applications. Lastly, approaches such as the intent-driven arbitration framework by Gonzalez et al.~\cite{gonzalez2024intent} leverage pose and sensor data to enable energy-efficient XR interactions by seamlessly switching between devices—such as smartphones, tablets, and smartwatches—based on user gaze and proximity.

In contrast to prior works, \emph{EgoTrigger} targets the usage of a less expensive, common sensing modality, audio from a microphone, in order to effectively gate image capture via pertinent audio cues (e.g., hand-object interaction audio cues). This enables energy savings specifically for vision-centric smart glasses capable of passive, intelligent sensing and human memory enhancement, without requiring continuous multimodal sensing or sacrificing too much downstream task performance.

\noindent\textbf{Audio-based Context Recognition.} Audio-based context recognition has become a pivotal component in wearable and egocentric computing, enabling devices to infer user activities and environmental cues with minimal energy consumption. Foundational tools like openSMILE~\cite{eyben2010opensmile}, AudioSet~\cite{gemmeke2017audio}, and YAMNet~\cite{yamnet2017} have been instrumental in classifying a wide range of audio events, from environmental sounds to speech and emotion detection. Models inspired by such tools are particularly valuable for resource-constrained devices due to their lightweight architectures and real-time processing capabilities. 

Recent advancements have focused on integrating audio with visual data to enhance context recognition, especially in egocentric settings. For instance, EPIC-Fusion~\cite{kazakos2019epic} employs mid-level fusion of audio, RGB, and optical flow modalities to improve action recognition in first-person videos. Listen to Look~\cite{gao2020listen}, a framework that leverages audio as a preview mechanism to reduce visual redundancy in action recognition tasks, enables efficient recognition by distilling video features from a single frame and its accompanying audio. Furthermore, foundation models like Qwen-Audio~\cite{chu2023qwen} have demonstrated strong performance across tasks such as audio classification, speech recognition, and emotion recognition, indicating their potential for robust audio understanding in diverse applications.

Prior works underscore the importance of audio as a low-power, informative modality for context recognition in wearable devices. By leveraging audio cues, systems like our proposed \emph{EgoTrigger} can efficiently manage sensor activation, reducing energy consumption while maintaining performance in applications such as human memory enhancement.

\noindent\textbf{Human Memory Enhancement from Egocentric Data.} Research on egocentric data, which most commonly involves visual data, has increasingly focused on memory augmentation, leveraging wearable devices to capture and retrieve personal experiences. Datasets like Ego4D~\cite{grauman2022ego4d} and an extension of its episodic memory benchmark QA-Ego4D~\cite{barmann2022did} provide extensive egocentric video data paired with question-answer annotations, supporting models that can answer queries such as "Where did I leave my keys?". Building on this, Encode-Store-Retrieve~\cite{shen2024encode} introduces a system that encodes egocentric videos into language representations stored in a vector database, enabling efficient natural language querying and outperforming traditional models in episodic recall tasks.

Computationally efficient approaches that leverage less expensive sensor modalities, such as IMU in the case of EgoDistill~\cite{tan2023egodistill} and audio in the case of Listen to Look~\cite{gao2020listen}, aim to reduce visual redundancy and enhance action recognition in egocentric videos. Additionally, emerging datasets such as EgoSchema~\cite{mangalam2023egoschema}, EgoTempo~\cite{plizzari2025omnia}, and EgoLife~\cite{yang2025egolife} aim to provide richer contextual information, including temporal dynamics and daily activity patterns, to support the development of more robust memory augmentation systems. These advancements highlight the potential of integrating multimodal egocentric data to create personalized, efficient, and context-aware memory aids.

In this paper, our proposed system, \emph{EgoTrigger}, advances the vision of all-day human memory enhancement by using lightweight audio classification capable of on-device usage to selectively trigger image capture during hand-object interaction moments, which in turn serve as key cues for episodic memory and subsequently reduce the need for continuous visual sensing while maintaining downstream performance for human memory enhancement.

%% file: sections/main/3-our_approach.tex
\section{Our Approach}
\label{sec:our_approach}

\subsection{EgoTrigger}

We hypothesize that inexpensive sensing modalities (e.g., audio from a microphone) will be readily available in future smart glasses and can be leveraged to effectively trigger capture of imagery from the camera. Prior research underscores the significance of hand-object interactions (HOIs) in shaping our memories, emphasizing that bodily engagement, particularly involving our hands, plays a crucial role in how we encode, recall, and structure key life experiences~\cite{iani2019embodied}. Our approach centers on utilizing audio cues characteristic of Hand-Object Interactions (HOI) as a trigger mechanism for egocentric data capture. The core component is a binary audio classifier, $f_{\theta}$, parameterized by $\theta$, trained to differentiate between audio segments containing HOI events ($C_1$) and those without HOI events ($C_0$). This classifier forms the foundation of our \emph{EgoTrigger} system. \emph{EgoTrigger} processes a continuous audio stream $X$ by applying $f_{\theta}$ within a sliding window framework. Specifically, it analyzes short, overlapping audio segments $x_t$ extracted from $X$ at time intervals corresponding to the window hop. Let $w_d$ be the window duration (e.g., 4 seconds) and $w_h$ be the hop size (e.g., 2 seconds).

$$
x_i = X[i \cdot w_h : i \cdot w_h + w_d]
$$

The classifier outputs the posterior probability:

$$
P(C_1 \mid x_i) = f_{\theta}(x_i)
$$

\emph{EgoTrigger} activates a designated function when this probability exceeds a predefined threshold $\tau$:

$$
P(C_1 \mid x_i) \ge \tau
$$

This provides contextually relevant activation based on audible user interactions. \emph{EgoTrigger} then translates the classifier's output into practical event triggers. When $P(C_1 \mid x_i) \ge \tau$, the system enters an active state. Two primary control strategies manage this state. The first employs a fixed \lq{OFF}\rq~duration, $T_{fixed}$ (e.g., 1 second). If the trigger activates at the start time $t_{start} = i \cdot w_h$ of window $i$, the system remains active for the interval:

$$
[t_{start},\ t_{start} + T_{fixed}]
$$

The second strategy uses hysteresis with distinct \lq{ON}\rq~and \lq{OFF}\rq~thresholds, $\tau_{on}$ and $\tau_{off}$ respectively, where $\tau_{on} > \tau_{off}$ (e.g., $\tau_{on}=0.8$, $\tau_{off}=0.7$). The system state $S_t \in \{\text{ON}, \text{OFF}\}$ at time $t$ is updated based on the classifier's output using the following hysteresis rule:

$$
S_t = 
\begin{cases}
\text{ON}, & \text{if } S_{t-1} = \text{OFF} \text{ and } P(C_1 \mid x_t) \ge \tau_{on} \\
\text{OFF}, & \text{if } S_{t-1} = \text{ON} \text{ and } P(C_1 \mid x_t) < \tau_{off} \\
S_{t-1}, & \text{otherwise}
\end{cases}
$$

This prevents rapid state changes due to minor probability fluctuations. Both strategies output time intervals:

$$
[t_{start},\ t_{stop}]
$$

defining periods of system activation for reference.

As a part of our evaluation in~\Cref{sec:egotrigger_evaluation}, we evaluate \emph{EgoTrigger}'s ability to differentiate between audio segments containing HOI events ($C_1$) and those without HOI events ($C_0$), additionally taking into account weighted results due to the inherent class imbalances to most egocentric datasets involving audio. We also analyze trigger robustness and false positives, including provision of a false positive rate. In~\Cref{sec:episodic_memory_task_evaluation}, we evaluate on the episodic memory task using both the fixed \lq{OFF}\rq~duration, $T_{fixed}$, strategy and the hysteresis with distinct \lq{ON}\rq~and \lq{OFF}\rq~thresholds strategy in contrast to continuous capture and fixed decimation baselines.

\subsection{Implementation}

The classifier $f_{\theta}$ was implemented using TensorFlow~\cite{developers2022tensorflow}, leveraging transfer learning from a pretrained YAMNet~\cite{yamnet2017} model. The final classification layer of YAMNet was replaced with a custom four-layer dense network, fine-tuned for binary hand-object interaction (HOI) detection ($C_1$ vs $C_0$). The classifier head consists of a sequence of fully connected layers with ReLU activations and dropout for regularization: 1024 $\rightarrow$ 256 $\rightarrow$ 384 $\rightarrow$ 192 $\rightarrow$ 384 $\rightarrow$ 2, with dropout rates of 0.15, 0.2, 0.25, and 0.2 respectively after each hidden layer.

At training time, input audio segments are pre-processed via a standardized pipeline that includes YAMNet as a base model to extract embeddings. This process includes conversion to mono, resampling to a target sampling rate $F_s = 16$ kHz, and amplitude normalization to the range $[-1.0,\ 1.0]$. The model outputs logits $z = [z_0, z_1]$, which are converted to class probabilities using the softmax function:

$$
P(C_k \mid x_i) = \frac{e^{z_k}}{e^{z_0} + e^{z_1}}, \quad k \in \{0, 1\}
$$

A decision threshold $\tau$ (e.g., $\tau = 0.4$) is then applied to $P(C_1 \mid x_i)$ to determine whether to activate the image capture trigger. We train \emph{EgoTrigger} using the binary cross-entropy loss:

$$
\mathcal{L} = -\left[ y \log f_{\theta}(x) + (1 - y) \log (1 - f_{\theta}(x)) \right]
$$

where $y \in \{0, 1\}$ is the ground-truth label indicating the presence ($C_1$) or absence ($C_0$) of a hand-object interaction, and $f_{\theta}(x)$ is the predicted probability of $C_1$.

In order to train \emph{EgoTrigger}, we used a subset of the Ego4DSounds dataset from~\cite{chen2024action2sound}. Ego4DSounds is derived from Ego4D~\cite{grauman2022ego4d}, an existing large-scale egocentric video dataset, and contains video clips spanning hundreds of different scenes and actions. Clips have a high action-audio correspondence and contain time-stamped narrations indicating actions performed by the camera-wearer, making them an excellent candidate to build training and test data for our particular classification task. We randomly sampled approximately 14,000 audio waveform files for training and approximately 2,300 samples for testing. Using Gemini 1.5 Pro, we additionally ran an audio understanding pass on all samples in order to annotate them with a label, based on both the audio and corresponding narration, indicating whether or not the audio sample corresponded to a hand-object interaction (HOI). We include further details regarding this process, including the prompt we utilized, in our supplementary materials. To address class imbalance in the Ego4DSounds dataset—where hand-object interaction (HOI) events can dominate—we applied a class weighting strategy. Class weights were computed from training set frequencies, resulting in a weight of $w_0 = 5.6577$ for the negative class ($C_0$) and $w_1 = 0.5485$ for the positive class ($C_1$). These weights were incorporated directly into the loss function.

For deployment, particularly targeting resource-constrained edge devices, the TensorFlow model was converted to the TensorFlow Lite (TFLite) format~\cite{david2021tensorflow}, more recently known as LiteRT. Default optimizations were applied to enhance efficiency. This resulted in a model size reduction of 86.23\%, bringing the model size from 23.60MB to 3.77MB. Additionally, it should be noted that YAMNet-based models and the LiteRT optimized variants are highly capable of being deployed on-device, with prior task benchmarks indicating an average latency on a mobile device using CPU / GPU being 12.29 milliseconds~\cite{mediapipe_audio_classifier}. This can be further optimized by the usage of known techniques, such as streaming convolutions~\cite{rybakov2020streaming} or quantization~\cite{krishna2023tinyml} that enables usage on specific hardware components such as the "low power island" of the AR1 Gen1~\cite{qualcommAR1}.

%% file: sections/main/4-HME-QA.tex
\section{HME-QA}
\label{sec:hme-qa}

\subsection{Motivation}

Recent advancements in egocentric datasets, notably Ego4D~\cite{grauman2022ego4d}, have catalyzed the development of task-specific extensions aimed at enhancing various aspects of video understanding. For instance, EgoSchema~\cite{mangalam2023egoschema} focuses on long-form video question answering, requiring models to comprehend extended temporal contexts. EgoTempo~\cite{plizzari2025omnia} emphasizes temporal reasoning capabilities, challenging models to integrate information across entire videos. Additionally, datasets like QA-Ego4D~\cite{barmann2022did} extend Ego4D's tasks by pairing egocentric videos with natural language questions and answers, facilitating research in natural language query (NLQ) understanding.

Despite the richness of these datasets, a significant limitation persists: the lack of comprehensive audio data and corresponding QA pairs. Audio cues can be crucial for tasks involving hand-object interactions (HOI), which are central to our approach. An analysis of the QA-Ego4D dataset reveals that nearly 40\% of its test split videos lack audio tracks, rendering them unusable for audio-centric research. This gap underscores the necessity for a curated dataset that ensures the presence of meaningful audio, thereby enabling more robust exploration of multimodal egocentric data.

\subsection{Dataset}

To address this limitation, we curated the HME-QA dataset, focusing on videos that encompass both high-quality audio and IMU modalities within the episodic memory benchmark of Ego4D. Our curation process began by filtering the Ego4D dataset to identify 82 full-length videos containing complete narrations and moments annotations. We then employed FFMPEG to verify the presence of actual audio tracks, resulting in a refined set of 50 valid videos. For these 50 videos, we utilized Gemini 2.0 Flash to generate initial question-answer pairs, emphasizing semantic relevance to the camera wearer's interactions. These AI-generated pairs underwent meticulous review by an expert annotator specializing in egocentric video datasets. The annotator discarded low-quality pairs and retained those deemed high-quality, culminating in a collection of 340 question-answer pairs.

The HME-QA dataset comprises 50 full-length Ego4D videos, on average 12 minutes in duration, with some extending up to 30 minutes. This dataset can serve as a valuable resource for advancing research in human memory enhancement through multimodal egocentric data, with a focus on ensuring high quality audio data is available in videos utilized for evaluation.

\begin{figure}[h!]
    \centering
    \includegraphics[width=\columnwidth]{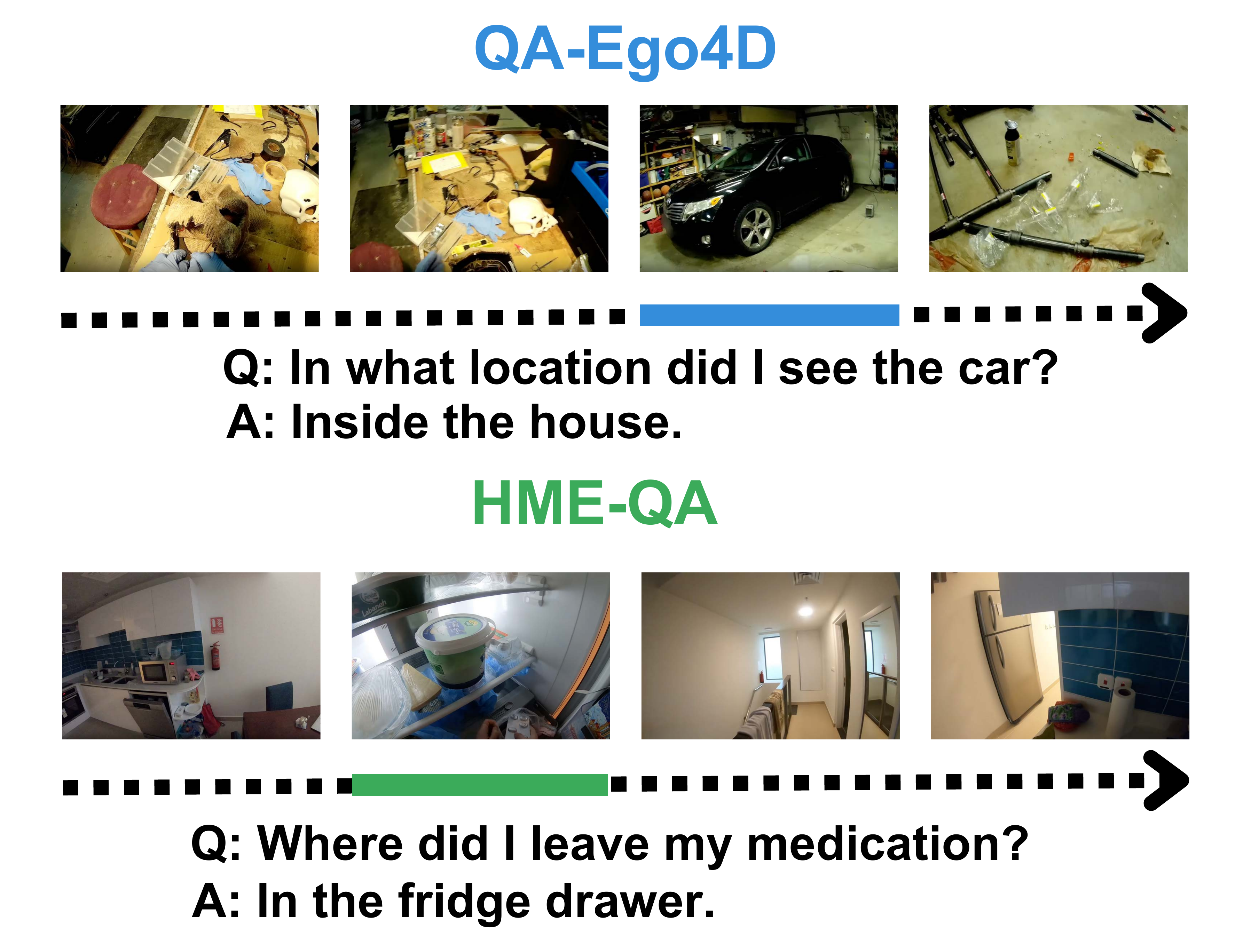}
    \vspace{-1.0em}
    \caption{\textbf{Sample QA Pairs from Datasets.} Example question-answer pairs from QA-Ego4D (top) and HME-QA (bottom), each paired with the relevant image sequence from which the answer is derived.}
    \label{fig:qa_datasets_comparison}
    \vspace{-1.0em}
\end{figure}

%% file: sections/main/5-experimental_setup.tex
\section{Experimental Setup}
\label{sec:experimental_setup}

We trained the EgoTrigger audio classifier using a batch size of 64, categorical cross-entropy loss, and the AdamW~\cite{kingma2014adam, loshchilov2017decoupled} optimizer with a learning rate $\eta = 3 \times 10^{-3}$ and weight decay $\lambda = 0.01$. Training was conducted for 50 epochs. Model performance on the Ego4DSounds test set was evaluated using standard classification metrics: precision ($\text{Pr}$), recall ($\text{Re}$), and F1-score, defined as $F_1 = \frac{2 \cdot \text{Pr} \cdot \text{Re}}{\text{Pr} + \text{Re}}$. We analyzed performance trade-offs across various decision thresholds $\tau$ through~\Cref{fig:c1_metrics_threshold}.

For evaluation, we performed the same filtering of videos for audio done in~\Cref{sec:hme-qa} for the QA-Ego4D dataset - taking us from an initial test set comprised of 1854 QA pairs across 166 unique videos to a final test set comprised of 1071 QA pairs across 104 unique videos. We paired these processed videos with their corresponding question-answer pairs and used a large language model (LLM) to score the correctness of predicted answers. Following prior work~\cite{plizzari2025omnia, khattak2024good, maaz2023video, qian2024easy}, we adopted an LLM-as-a-judge evaluation protocol. Specifically, we employed Gemini models with a fixed seed of 1337 and a temperature of 0.2 to ensure consistent results. For generating answers from processed videos and questions, we utilized Gemini 1.5 Pro, while for judging we utilized Gemini 1.5 Pro in a separate pass. To isolate attention and ensure precise evaluation, each prompt contained a single question paired with its corresponding video segment. Further details, including the exact prompts used, are provided in the supplementary materials.

%% file: sections/main/6-evaluation.tex
\section{Evaluation}

\subsection{Performance Evaluation of EgoTrigger}
\label{sec:egotrigger_evaluation}

We evaluate \emph{EgoTrigger}'s ability to detect hand-object interaction (HOI) events by comparing several strategies to handle the imbalanced Ego4DSounds test set, with detailed results shown in \cref{tab:hoi_classification_with_support}. Our baseline approach using class weights achieves strong performance on the positive majority class ($C_1$), attaining an F1-score of 0.91. However, performance on the negative minority class ($C_0$) is poor, with an F1-score of only 0.15. To address this, we implemented an oversampling strategy using SMOTE. This approach dramatically improves performance for the minority class, boosting the F1-score for $C_0$ from 0.15 to 0.54, while maintaining excellent performance on $C_1$ (0.90 F1-score). This results in a superior overall model, increasing the weighted average F1-score from 0.83 to 0.86. For completeness, we also evaluated random undersampling, which improved the $C_0$ F1-score to 0.47 but degraded the $C_1$ F1-score significantly, lowering the overall weighted average. Given these results, we will adopt the model trained with SMOTE for our subsequent experiments.

\begin{table}[h!]
\centering
\caption{\textbf{HOI Classification Performance.} Evaluation of binary hand-object interaction (HOI) classification on the test set using three different training strategies. The support numbers remain constant as resampling is only applied to training data.}
\label{tab:hoi_classification_with_support}
    \begin{tabularx}{\linewidth}{@{} l >{\raggedright\arraybackslash}X c c c c @{}} 
    \toprule
    \textbf{Strategy} & \textbf{Class} & \textbf{Prec.} & \textbf{Recall} & \textbf{F1} & \textbf{Supp.} \\
    \midrule
    \multirow{3}{*}{\begin{tabular}[c]{@{}l@{}}Class Weights\\ (Original)\end{tabular}} 
    & $C_0$ (No HOI)   & 0.17 & 0.14 & 0.15 & 251  \\
    & $C_1$ (HOI)      & 0.90 & 0.92 & 0.91 & 2095 \\
    & Weighted Avg   & 0.82 & 0.84 & 0.83 & 2346 \\
    \midrule
    \multirow{3}{*}{\begin{tabular}[c]{@{}l@{}}Oversampling\\ (SMOTE)\end{tabular}} 
    & $C_0$ (No HOI)   & 0.46 & 0.65 & 0.54 & 251  \\
    & $C_1$ (HOI)      & 0.93 & 0.87 & 0.90 & 2095 \\
    & Weighted Avg   & 0.88 & 0.85 & 0.86 & 2346 \\
    \midrule
    \multirow{3}{*}{\begin{tabular}[c]{@{}l@{}}Undersampling\\ (Random)\end{tabular}}
    & $C_0$ (No HOI)   & 0.35 & 0.70 & 0.47 & 251  \\
    & $C_1$ (HOI)      & 0.94 & 0.74 & 0.83 & 2095 \\
    & Weighted Avg   & 0.88 & 0.73 & 0.79 & 2346 \\
    \bottomrule
    \end{tabularx}
\end{table}

\begin{figure}[h!]
    \centering
    \includegraphics[width=\columnwidth]{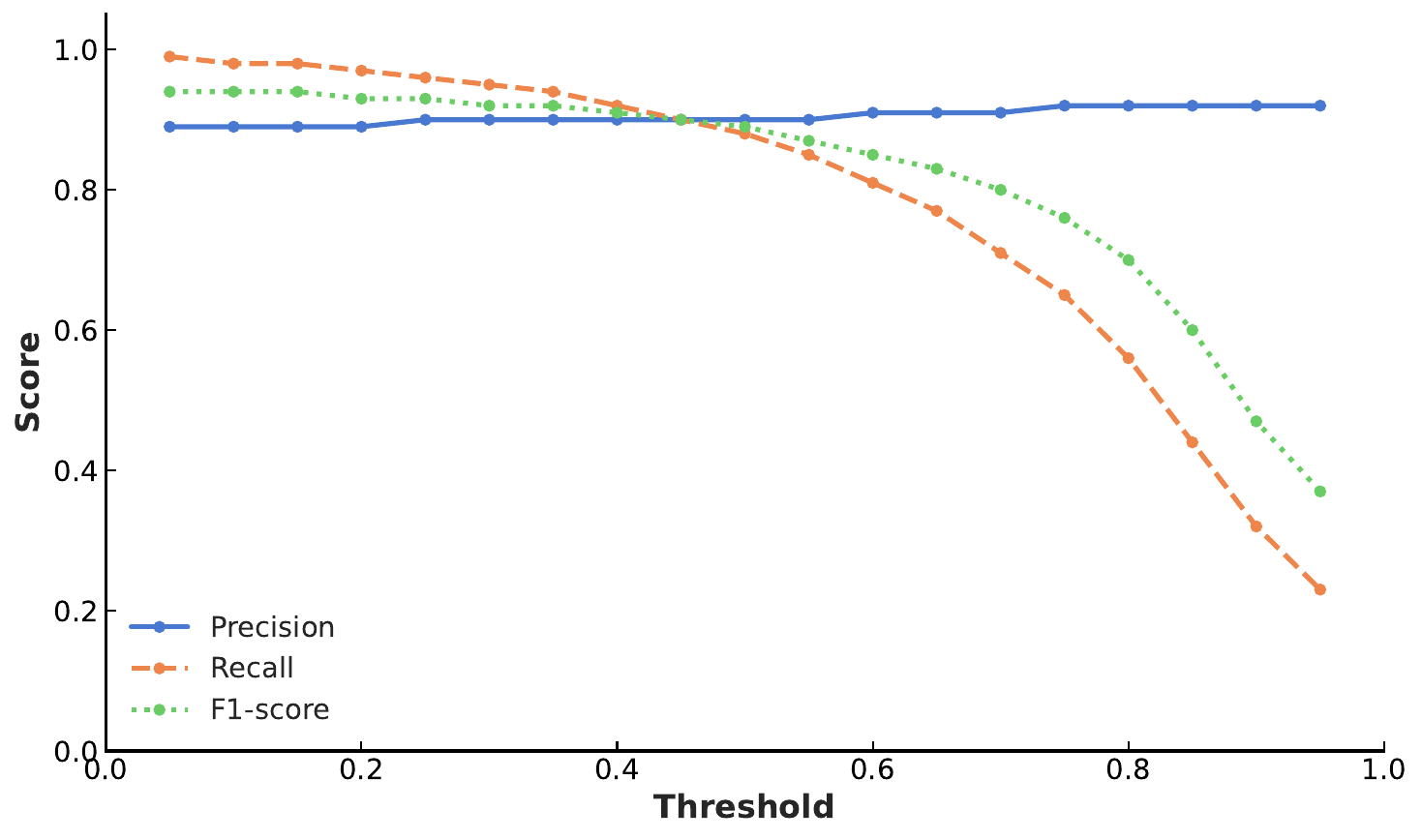}
    \vspace{-1.0em}
    \caption{\textbf{$\mathbf{C_1}$ Metrics vs. Threshold.} Precision, recall, and F1-score for class $\mathbf{C_1}$ as a function of decision threshold. Precision remains high across thresholds, while recall decreases more rapidly. F1-score peaks near the region where precision and recall are balanced, guiding threshold selection for optimal tradeoff.}
    \label{fig:c1_metrics_threshold}
    \vspace{-1.0em}
\end{figure}

To analyze the effect of the decision threshold $\tau$, we report precision, recall, and F1-score for class $C_1$ across a range of thresholds in \cref{fig:c1_metrics_threshold}. Results indicate that EgoTrigger maintains robust F1-scores ($\ge 0.9$) for a wide range of thresholds ($\tau \le 0.5$), highlighting the reliability of the classifier across deployment settings. As $\tau$ increases, the model favors precision over recall, enabling applications to trade off between false positives and false negatives depending on the energy constraints and tolerance for missed events.

To further assess the robustness and real-world viability of our HOI trigger, we conducted a series of experiments to quantify its performance against non-HOI sounds, additive noise, and in continuous, uncurated video streams. The results are summarized in \cref{tab:trigger_robustness}.

First, we evaluated the trigger's specificity by testing it against two datasets that should not contain HOI events: the standard ESC-50 environmental sound dataset and a curated (rather than randomly sampled) subset of Ego4DSounds including human speech. Our trained model demonstrates high specificity, with a False Positive Rate (FPR) of only 7.1\% on the diverse environmental sounds of ESC-50 and a very low 2.3\% on a version of the Ego4DSounds test set that includes speech. This indicates that \emph{EgoTrigger} is well-tuned to HOI sounds and is not trivially triggered by general ambient noise or conversation.

Second, we tested the trigger's resilience to noise by adding white noise to our Ego4DSounds test set at various Signal-to-Noise Ratios (SNRs). While performance degrades in noisier conditions, the model maintains a strong weighted F1-score of 0.78 even at a higher noise level ($\sigma^2=0.1$).

Finally, to evaluate real-world performance, we tested the system on five full-length videos from our HME-QA dataset, manually identifying false triggers unrelated to human-object interactions. The system generated an average of 2.4 false positives per minute (FPPM). This rate allows the system to remain in standby mode most of the time while avoiding continuous recording, making it viable for practical deployment despite remaining challenges with noise robustness.

\begin{table}[h!]
\centering
\caption{\textbf{Trigger Robustness and False Positive Analysis.} False Positive Rate (FPR) is shown on non-HOI datasets, alongside F1-score degradation with additive noise and the in-the-wild False Positives Per Minute (FPPM) rate.}
\label{tab:trigger_robustness}
    \begin{tabular}{@{}ll@{}}
    \toprule
    \textbf{Experiment / Condition} & \textbf{Value} \\
    \midrule
    \textbf{Specificity (FPR)} & \\
    \quad ESC-50 Environmental Sounds & 7.1\% \\
    \quad Ego4DSounds (Non-HOI Test Clips) & 2.3\% \\
    \midrule
    \textbf{Robustness to Noise (F1-score)} & \\
    \quad Clean ($\sigma^2$=0) & 0.86 \\
    \quad Low Noise ($\sigma^2=0.01$) & 0.78 \\
    \quad High Noise ($\sigma^2=0.1$) & 0.59 \\
    \midrule
    \textbf{In-the-Wild Perf. (FPPM)} & \\
    \quad 5 HME-QA Videos (avg. 12 min) & 2.4 \\
    \bottomrule
    \end{tabular}
\end{table}

\subsection{Performance on Egocentric QA Datasets}
\label{sec:episodic_memory_task_evaluation}

To simulate \emph{EgoTrigger}-based image capture on input video, we first processed the input audio waveform using a sliding window approach with a window duration $w_d = 4$ seconds and a hop size $w_h = 2$ seconds. EgoTrigger includes an offline operation mode that processes pre-recorded video by simulating trigger behavior using predicted timestamps for when image capture would be activated or suppressed. Using the predicted activation intervals, we generate timestamped JSON files for each video. These are then used to modify the original videos via FFMPEG by blacking out all frames that would not have been captured by EgoTrigger.

\begin{figure*}[t!]
    \centering
    \includegraphics[width=1\textwidth]{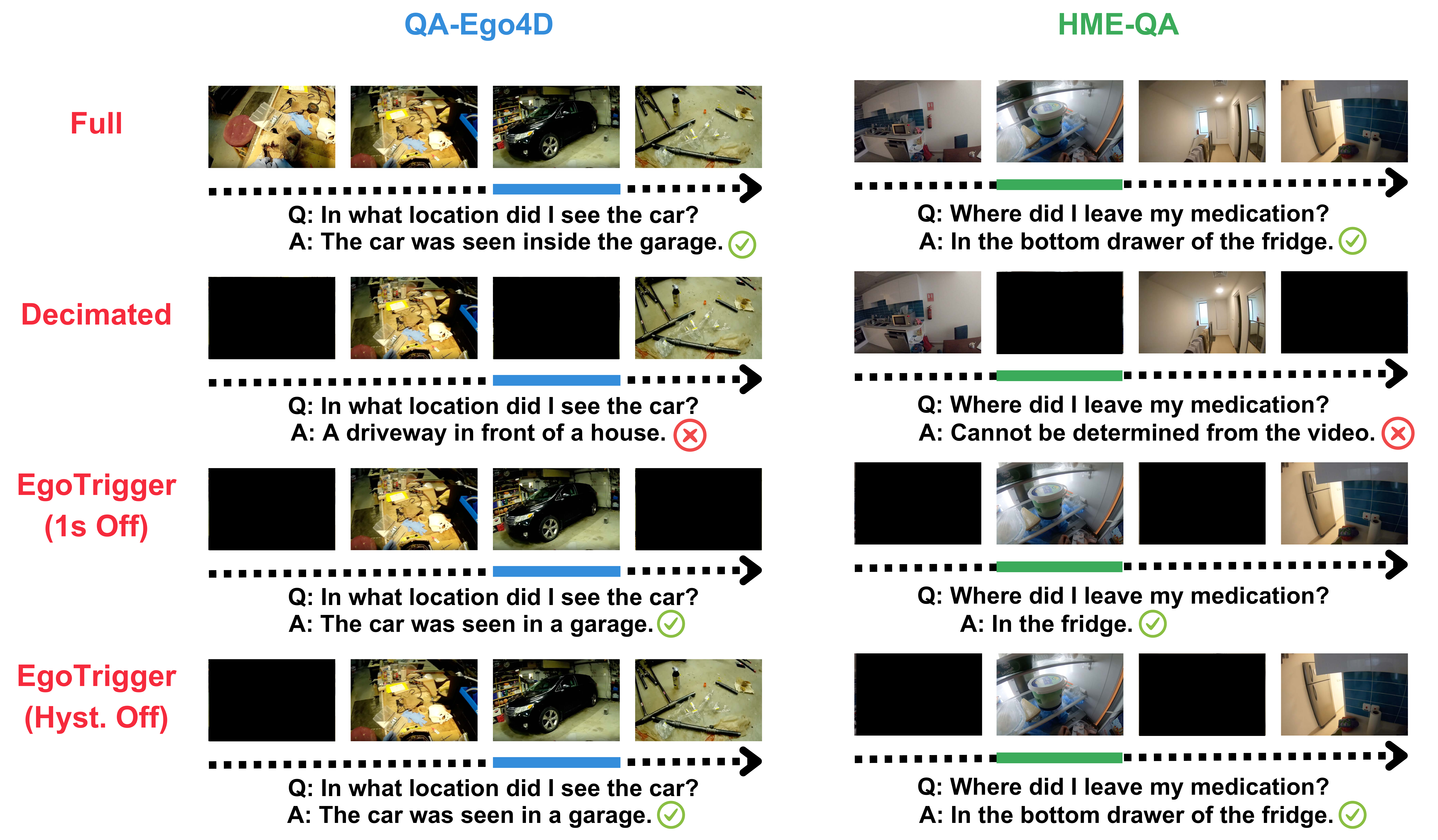}
    \vspace{-2em}
    \caption{\textbf{EgoTrigger QA Results.} Sample question-answer outcomes on QA-Ego4D (left) and HME-QA (right) using different video sampling strategies. EgoTrigger variants preserve key visual content necessary for correct answers while significantly reducing the number of frames, compared to full or naively decimated baselines.}
    \label{fig:egotrigger_qa_results_example}
    \vspace{-0.4cm}
\end{figure*}

We evaluated four approaches—continuous full capture (\textbf{Full}), naive decimation (\textbf{Decimated}), EgoTrigger with fixed-duration triggering (\textbf{ET-1s}), and EgoTrigger with hysteresis-based control (\textbf{ET-Hyst.})—on both the HME-QA and QA-Ego4D datasets. Naive decimation—uniformly dropping frames at fixed intervals—can reduce bitrate and energy consumption but often sacrifices temporal coherence, leading to degraded performance on downstream video understanding tasks~\cite{herglotz2023video}. Our naive decimation approach involves processing a video such that frames are only captured once every five seconds, effectively operating at 0.2 FPS. On HME-QA, the Full baseline achieved the highest accuracy at 77.3\%. Both EgoTrigger variants maintained more comparable performance, with ET-1s and ET-Hyst. achieving 75.7\% and 74.7\% accuracy, respectively. In contrast, decimation dropped performance to 71.9\%. On QA-Ego4D, the Full approach reached 41.08\% accuracy, followed closely by ET-1s (40.1\%) and ET-Hyst. (39.2\%), while decimation again underperformed at 34.6\%. These results indicate that EgoTrigger can significantly reduce visual data while preserving QA performance close to the full capture baseline. Qualitative examples are shown in~\Cref{fig:egotrigger_qa_results_example}.

\subsection{Performance versus Energy Efficiency}
\label{sec:approach_analysis_performance_versus_energy}

Tables~\ref{tab:hmeqa_acc_red_br} and~\ref{tab:qaego4d_acc_red_br} present a comparative analysis of four image capture strategies across two egocentric QA datasets: HME-QA and QA-Ego4D. These include continuous full capture (\textbf{Full}), uniform frame decimation (\textbf{Decimated}), EgoTrigger with a fixed 1-second OFF duration (\textbf{ET-1s}), and EgoTrigger with a hysteresis-based OFF trigger (\textbf{ET-Hyst.}). \Cref{fig:accuracy_versus_reduction_pareto} contains a visualization of the accuracy versus frame reduction trade-off.

\begin{table}[h!]
\centering
\label{tab:qa_correctness_combined}
\caption{\textbf{Accuracy (\%) across Datasets.} Comparison of correctness (rating $\geq 3$) scores across approaches on HME-QA ($N=340$) and QA-Ego4D ($N=1071$). The best result is in bold, while the second best result is underlined.}
\begin{tabular}{@{}lcc@{}}
\toprule
\textbf{Approach} & \textbf{HME-QA} & \textbf{QA-Ego4D} \\
\midrule
Full        & \textbf{77.3} & \textbf{41.08} \\
Decimated   & 71.9 & 34.6 \\
ET-1s       & \underline{75.7} & \underline{40.1} \\
ET-Hyst.    & 74.7 & 39.2 \\
\bottomrule
\end{tabular}
\end{table}

\begin{table}[t!]
\centering
\caption{\textbf{Performance and Efficiency on HME-QA.} Comparison of accuracy, frame reduction, and estimated bitrate across baselines and EgoTrigger variants on the HME-QA dataset.}
\resizebox{\columnwidth}{!}{%
\begin{tabular}{@{}lccc@{}}
\toprule
\textbf{Approach} & \textbf{Correct (\%)} & \textbf{Frames Reduced (\%)} & \textbf{Est. Bitrate (Mbps)} \\
\midrule
Full        & 77.3 & --    & 5.47 \\
Decimated   & 71.9 & 79.48 & 1.13 \\
ET-1s       & 75.7 & 54.39 & 2.50 \\
ET-Hyst.    & 74.7 & 27.88 & 3.96 \\
\bottomrule
\end{tabular}%
}
\label{tab:hmeqa_acc_red_br}
\end{table}

\begin{table}[t!]
\centering
\caption{\textbf{Performance and Efficiency on QA-Ego4D.} Comparison of accuracy, frame reduction, and estimated bitrate across baselines and EgoTrigger variants on the QA-Ego4D dataset.}
\resizebox{\columnwidth}{!}{%
\begin{tabular}{@{}lccc@{}}
\toprule
\textbf{Approach} & \textbf{Correct (\%)} & \textbf{Frames Reduced (\%)} & \textbf{Est. Bitrate (Mbps)} \\
\midrule
Full        & 41.08 & --    & 1.31 \\
Decimated   & 34.6  & 79.41 & 0.27 \\
ET-1s       & 40.1  & 54.28 & 0.60 \\
ET-Hyst.    & 39.2  & 23.22 & 1.01 \\
\bottomrule
\end{tabular}%
}
\label{tab:qaego4d_acc_red_br}
\end{table}

Across both datasets, EgoTrigger variants consistently reduce frame usage by a substantial margin while maintaining QA accuracy close to that of full capture, in contrast to the decimated baseline. For HME-QA, ET-1s achieves a 54.39\% reduction in frames and only a 1.7 percentage point drop in accuracy compared to Full. Similarly, on QA-Ego4D, ET-1s reduces frame count by 54.28\% with only a 0.98 percentage point drop in accuracy. In contrast, decimation saves the most frames (79.5\%) but at a greater accuracy cost, especially on QA-Ego4D where it drops by nearly 6.48 percentage points.

These reductions can translate directly to bitrate savings. ET-1s approximately halves the bitrate on both datasets (HME-QA: 2.50 Mbps vs. 5.47 Mbps Full; QA-Ego4D: 0.60 Mbps vs. 1.31 Mbps Full). We inferred these values using black frame durations calculated via \texttt{FFMPEG} and video file sizes. Black frames replace periods where the trigger was off, and thus serve as a reliable estimate of when camera capture can be off and effective data transmission avoided.

%% file: sections/main/7-discussion.tex
\section{Discussion}

Our evaluation demonstrates the potential of \emph{EgoTrigger} as an energy-efficient strategy for capturing visually relevant moments for memory enhancement, triggered by audio cues. We now discuss the implications of our findings, framed around the specific evaluation areas, followed by limitations and future directions.

\subsection{Performance Evaluation of EgoTrigger}
\label{sec:discussion_performance_eval_egotrigger}
The core technical feasibility of our approach hinges on reliably detecting Hand-Object Interactions (HOIs) from audio using a lightweight model suitable for on-device deployment. Our evaluation (\Cref{sec:egotrigger_evaluation}) confirmed that the fine-tuned YAMNet-based classifier can effectively identify characteristic HOI sounds within egocentric audio streams, even amidst significant class imbalance (\cref{tab:hoi_classification_with_support}). This validates the fundamental premise that salient audio cues can serve as effective triggers for visual capture. Furthermore, the classifier's robustness across various decision thresholds (\cref{fig:c1_metrics_threshold}) suggests practical deployability, allowing system designers to tune the trade-off between trigger sensitivity (recall) and specificity (precision) based on application needs and energy constraints. While classification performance on non-HOI sounds ($C_0$) was lower, the downstream task results indicate that the positive HOI signal is sufficiently discriminative for the intended memory enhancement application.

\subsection{Performance on Egocentric QA Datasets}
\label{sec:discussion_performance_egocentric_qa}

The key utility of EgoTrigger lies in its ability to support downstream applications while reducing resource consumption. Our evaluation on the HME-QA and QA-Ego4D datasets (\Cref{sec:episodic_memory_task_evaluation}) reveals a crucial insight: selectively capturing visual frames around detected HOIs can preserve the essential information required to answer challenging episodic memory questions. Both EgoTrigger variants maintained question-answering correctness remarkably close to that achieved using the full, continuous video stream (\Cref{tab:qa_correctness_combined}). This outcome contrasts sharply with naive temporal decimation, which, despite potentially higher frame reduction, incurred a more significant performance penalty. This suggests that the context-aware nature of EgoTrigger—focusing capture on moments of interaction—is substantially more effective at retaining task-relevant visual details than uniform subsampling. These findings support the hypothesis that HOIs often coincide with moments of high visual salience critical for memory encoding and retrieval, at least for the interaction-centric queries common in egocentric memory tasks. The evaluation also underscores the value of curated datasets like HME-QA, which guarantee audio availability for multimodal analysis (\Cref{sec:hme-qa}).

To further analyze these post-trigger recognition characteristics, it is important to clarify how intermittent video is processed. Our method does not create disjointed clips; rather, it blacks out frames from the original continuous video stream (\Cref{sec:episodic_memory_task_evaluation}). This means the downstream Large Multimodal Model (LMM) receives a video that retains its original duration and temporal structure, ensuring that the timing and distance between triggered events remain coherent. Our approach is effectively "training-free" for the downstream task; no architectural changes or fine-tuning were performed on the Gemini 1.5 Pro model. The minimal performance degradation observed between \emph{EgoTrigger} and the "Full" capture baseline (e.g., a $<$2\% drop for ET-1s on HME-QA) can be seen as the empirical answer to how performance fundamentally differs. We attribute this robustness to two factors: 1) the inherent capability of large foundation models, trained on vast and varied web-scale data, to perform reasoning with incomplete or sparse information, and 2) our system's success in preserving the most information-dense segments (HOIs) critical for this specific episodic memory task.

\subsection{Performance versus Energy-Efficiency}
\label{sec:discussion_performance_versus_energy-efficiency}
Achieving all-day operation in form-factor constrained smart glasses necessitates aggressive energy management. The substantial frame reduction achieved by \emph{EgoTrigger} (e.g., over 54\% reduction for ET-1s, see \cref{tab:hmeqa_acc_red_br}, \cref{tab:qaego4d_acc_red_br}) directly translates to potential energy savings by minimizing the active time of the most power-hungry components in the visual pipeline: the camera sensor, image signal processor (ISP), compression hardware, and wireless transmitter. Given the significant power draw associated with continuous high-resolution video streaming (estimated at ~770\,mW in our analysis, rapidly depleting typical smart glass batteries), the savings realized by gating these components during periods deemed less critical by the audio trigger can drastically extend device longevity.

However, a complete energy analysis must also consider the cost of the trigger mechanism itself. While \emph{EgoTrigger} saves power by deactivating the visual pipeline, it requires the audio pipeline (microphone, front-end processing) and the audio classifier to run continuously or near-continuously. The energy consumption associated with this – particularly the on-device model's inference cost – represents an overhead compared to a completely idle system. Quantifying this overhead precisely requires measurements on target hardware, as it depends heavily on the chosen platform (e.g., specific chipset capabilities like low-power DSPs or AI accelerators), the final model architecture, and the level of optimization applied (potentially involving frameworks like TFLite Micro for highly constrained devices).

\begin{figure}[h!]
    \centering
    \includegraphics[width=\columnwidth]{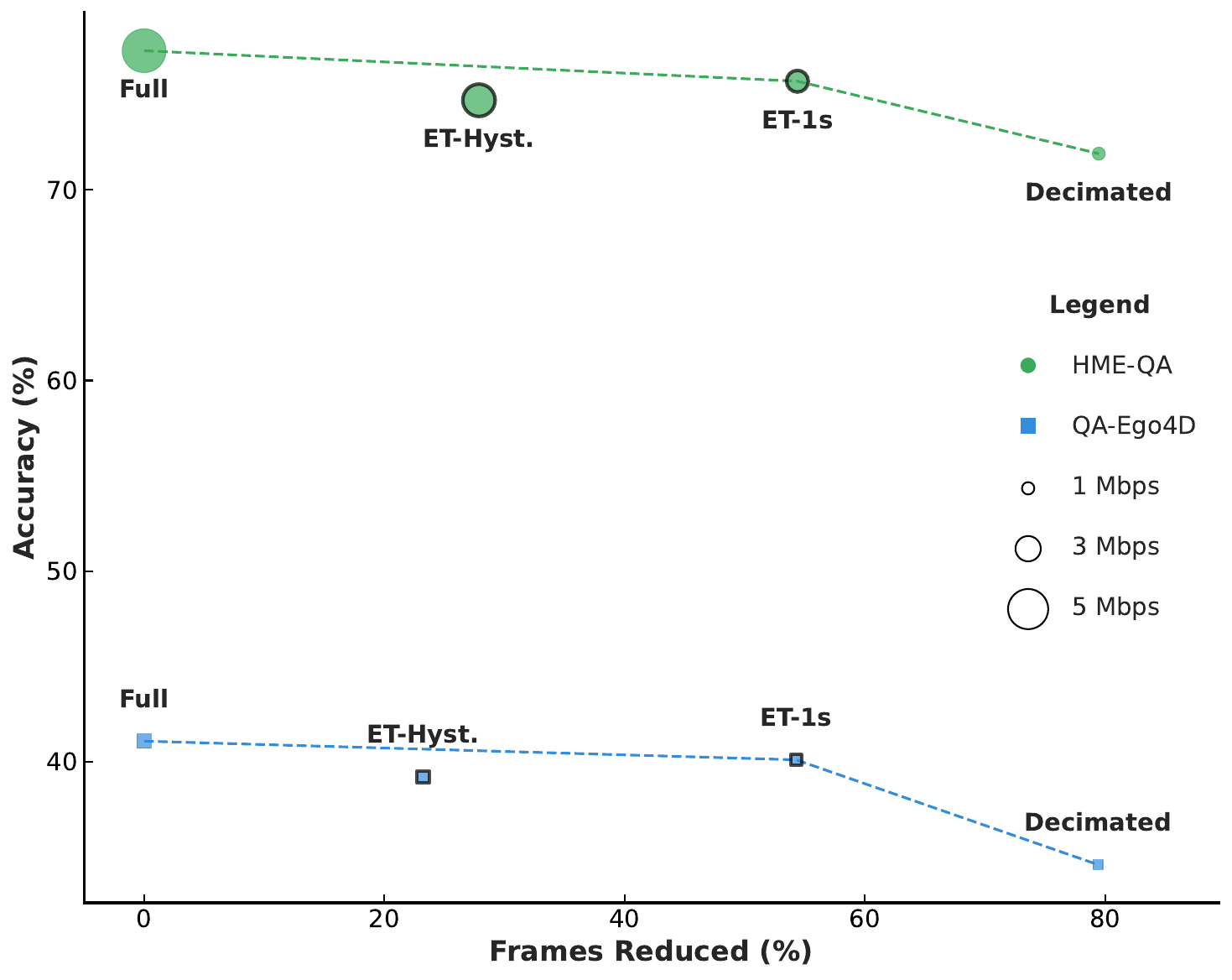}
    \vspace{-1.0em}
    \caption{\textbf{Accuracy vs. Frame Reduction Tradeoff.} \emph{EgoTrigger} variants (ET-1s and ET-Hyst.) lie close to the Pareto frontier, balancing QA accuracy with reduced frame count. Results are shown for both HME-QA and QA-Ego4D datasets, with dataset indicated by marker shape (circles for HME-QA, squares for QA-Ego4D) and estimated bitrate reflected by marker size. Outlined markers highlight our proposed methods. Variant ET-1s in particular reduces frame usage and subsequently bitrate while preserving accuracy close to full-frame baselines.}
    \label{fig:accuracy_versus_reduction_pareto}
    \vspace{-1.0em}
\end{figure}

Despite this inference cost, the energy savings from selectively deactivating the significantly more power-intensive visual pipeline are expected to yield substantial net power reduction compared to continuous visual capture. The favorable position of \emph{EgoTrigger} variants on the accuracy-versus-reduction Pareto frontier (\Cref{fig:accuracy_versus_reduction_pareto}) indicates a promising balance. Furthermore, the different triggering strategies (ET-1s vs. ET-Hyst.) offer tunability, enabling system designers to adapt the balance between energy conservation and potential information loss based on specific device limitations and user requirements, paving the way towards practical all-day assistive smart glasses.

\subsection{Empirical Power Consumption Analysis}

To empirically validate energy savings, which is infeasible on closed-platform commercial smart glasses, we constructed a simple hardware prototype. This system consists of a Raspberry Pi 5, a USB camera, and a USB microphone, with total power measured using a high-precision USB multimeter. Using this hardware, we measured the total system power draw across several key operating states: an idle baseline (0 FPS), our \emph{EgoTrigger} (ET-1s) approach, a naive decimation baseline, and two high-power baselines representing continuous capture and continuous capture with Wi-Fi transmission.

The empirical results in \cref{tab:power_measurements} demonstrate tangible power savings. Our prototype establishes a 2.37\,W idle power floor. From this, the average power for \emph{EgoTrigger} is 4.11\,W, representing a \textbf{17.3\% power reduction} compared to the 4.97\,W continuous capture baseline. This shows substantial savings from selective sensing alone. Furthermore, by reducing captured data by 54\%, \emph{EgoTrigger} avoids the high overhead of continuous wireless transmission (5.51\,W), a critical factor for enabling all-day battery life. While this prototype provides a valuable comparison, we note that a commercial product would use a highly optimized embedded system (e.g., an MCU or ASIC) with a much lower absolute power draw. However, we expect the \textit{relative} energy savings demonstrated by our selective triggering strategy to be comparable in such a specialized system.

\begin{table}[htbp]
\centering
\caption{\textbf{Empirical Power Analysis.} Measured power (Watts) of the Raspberry Pi 5 prototype, showing total system draw for each operating state and the contribution of its core components.}
\label{tab:power_measurements}
    \begin{tabular}{@{}lc@{}}
    \toprule
    \textbf{Operating Condition / Component} & \textbf{Power (W)} \\
    \midrule
    \textbf{System Idle (0 FPS Capture)} & \textbf{2.37} \\
    \quad \textit{Raspberry Pi 5 System (Idle)} & \textit{2.20} \\
    \quad \textit{Microphone (Connected)} & \textit{0.17} \\
    \midrule
    \textbf{Baseline (Continuous Capture)} & \textbf{4.97} \\
    \quad \textit{Raspberry Pi 5 System (Processing)} & \textit{4.20} \\
    \quad \textit{Microphone (Always-On)} & \textit{0.17} \\
    \quad \textit{Camera (Always-On)} & \textit{0.60} \\
    \midrule
    \textbf{Baseline + Continuous Wi-Fi Tx} & \textbf{5.51} \\
    \quad \textit{Raspberry Pi 5 System (Proc. + Tx)} & \textit{4.74} \\
    \quad \textit{Microphone (Always-On)} & \textit{0.17} \\
    \quad \textit{Camera (Always-On)} & \textit{0.60} \\
    \midrule
    \textbf{Decimated (0.2 FPS)} & \textbf{3.02} \\
    \quad \textit{Raspberry Pi 5 System (Processing)} & \textit{2.60} \\
    \quad \textit{Microphone (Always-On)} & \textit{0.17} \\
    \quad \textit{Camera (Gated, Avg.)} & \textit{0.25} \\
    \midrule
    \textbf{EgoTrigger (ET-1s, Avg. Power)} & \textbf{4.11} \\
    \quad \textit{Raspberry Pi 5 System (Processing)} & \textit{3.50} \\ 
    \quad \textit{Microphone (Always-On)} & \textit{0.17} \\
    \quad \textit{Camera (Gated, Avg.)} & \textit{0.44} \\
    \bottomrule
    \end{tabular}
\end{table}

%% file: sections/main/8-limitations_and_future_work.tex
\section{Limitations and Future Work}

Despite the promising results, our work highlights several limitations and opens avenues for future investigation:

\textit{Hardware Validation, Optimization, and Power Modeling:} Our current evaluation relied on offline processing of datasets and high-level energy estimations based on component datasheets and a simple hardware prototype. Validating \emph{EgoTrigger} on real smart glasses hardware is essential to measure true end-to-end performance, latency, and power consumption under real-world conditions. This includes accurately quantifying the on-device model's inference cost. Furthermore, deploying the machine learning model effectively onto resource-constrained smart glasses hardware will necessitate significant optimization efforts, potentially leveraging frameworks like TensorFlow Lite Micro (TFLM) which are designed for such environments, going beyond the TFLite conversion explored here. Crucially, such optimizations often involve trade-offs, and the potential impact on model accuracy or other performance metrics resulting from these necessary optimization steps has not been quantified in this study.

\textit{Trigger Robustness and Scope:} The current system depends exclusively on audible Hand-Object Interactions (HOIs) as triggers. This limits its applicability in scenarios involving silent actions (e.g., picking up a soft object), purely visual memories (e.g., observing something without interaction), or environments with high levels of masking noise or unfamiliar sounds that could affect trigger accuracy. Future work should explore integrating complementary low-power sensor modalities. This could include onboard Inertial Measurement Unit (IMU) data from the glasses to detect motion patterns associated with interactions, or even fusing data streams from other wearable devices the user might be wearing concurrently, such as smartwatches, fitness trackers, or potentially IMUs integrated into footwear. Combining these signals could help detect non-auditory interactions, disambiguate sounds, and improve overall trigger robustness. Additionally, exploring more advanced or personalized on-device audio models could enhance robustness to noise and user-specific interaction sounds. Furthermore, future work could explore learning audio representations from unlabeled video, inspired by seminal works like SoundNet~\cite{aytar2016soundnet}. While our current system relies on a pre-trained YAMNet and supervised fine-tuning, a cross-modal self-supervised approach could learn more robust features by leveraging the natural synchronization between visual data and audio. For example, by using a visual model to detect hand-object interactions in unlabeled egocentric videos, we could automatically generate weak labels to train an audio model to recognize the corresponding HOI sounds. This could significantly enhance the trigger's robustness across diverse environments and reduce the dependency on manually curated audio datasets, representing a promising direction for scalable and adaptive on-device sensing.

\textit{Evaluation Scope:} While the LLM-based Question-Answering evaluation provides valuable insights into downstream task performance, it should be complemented by real-time user studies. Assessing the practical usability and effectiveness of \emph{EgoTrigger} in diverse, real-world scenarios with human participants is crucial. Exploring integration with active user interaction models, such as allowing users to explicitly query their memories or providing feedback via a display, would also be valuable future directions.

\textit{Privacy Considerations:} As with any system involving continuous environmental sensing, particularly audio, careful consideration of user privacy is paramount. Deploying such technology requires robust safeguards, transparent user controls regarding data capture and processing, and adherence to ethical guidelines to ensure user trust and acceptance. \emph{EgoTrigger} shows promise in this direction by gating the transmission of sensor data unless required by a specific experience - in this case, particular human-object interactions for human memory enhancement.

Addressing these points will be crucial for realizing the full potential of passive, intelligent multimodal sensing systems for applications such as human memory enhancement, through wearable devices such as smart glasses.

%% file: sections/main/9-conclusion.tex
\section{Conclusion}

Enabling all-day human memory enhancement via smart glasses demands solutions to the significant energy cost associated with continuous camera operation. This paper introduced \emph{EgoTrigger}, an approach leveraging lightweight audio classification of hand-object interactions (HOIs) to intelligently trigger selective image capture. Our evaluations demonstrated that \emph{EgoTrigger} can reduce the number of captured visual frames by an average of 54\% while maintaining comparable performance on demanding episodic memory question-answering tasks, significantly outperforming naive decimation strategies. By selectively activating power-hungry visual sensors primarily when contextually relevant audio cues indicate moments likely critical for memory encoding, \emph{EgoTrigger} offers a practical pathway towards realizing energy-efficient smart glasses capable of passive, all-day sensing. This supports applications such as helping users recall daily activities or object locations. The proposed context-aware triggering strategy represents a meaningful step forward for assistive wearable computing, highlighting the potential of using low-power sensor modalities to intelligently manage high-power ones.

%% file: sections/main/10-acknowledgments.tex
\section{Acknowledgments}

Akshay Paruchuri conducted this research during an internship under the mentorship of Ishan Chatterjee at Google AR in Seattle, WA, USA. The authors thank the following colleagues at Google for insightful discussions around multimodal sensing and energy efficient computing: Zhihan Zhang, Jake Garrison, Shamik Ganguly, Zixuan Qu, and Anish Prabhu.

%% file: sections/supp/0-all_supp.tex
\appendix

\section{Generation of Initial QA Pairs for HME-QA}
\label{sec:appendix_hmeqa_generation}

To address the need for an egocentric QA dataset with reliable audio for memory enhancement research, we first selected and meticulously verified 50 full-length videos containing audio and complete narrations from the Ego4D episodic memory benchmark~\cite{grauman2022ego4d}. For these videos, we utilized the Gemini 2.0 Flash model to automatically generate an initial, large set of candidate question-answer pairs. A temperature of 0.7 and fixed seed of 1337 was utilized. The model was tasked with processing the video content and associated metadata (narrations, moments annotations), guided by a prompt instructing it to create QA pairs focused on the semantic content of the camera wearer's interactions within the scene. The specific prompt template employed for this initial generation phase is detailed below.

Recognizing that LLM-generated content requires careful validation, these candidate QA pairs were then subjected to a rigorous human curation process. An expert annotator, specializing in egocentric video data, meticulously reviewed each AI-generated pair against the source video content. Pairs that were factually inaccurate, irrelevant, ambiguous, or otherwise failed to meet the quality requirements for memory-related interaction queries were discarded. This essential human-in-the-loop filtering resulted in the final HME-QA dataset, comprising 340 high-quality, validated QA pairs across the 50 videos.

\section{Generation of Hand-Object Interaction (HOI) Labels for Ego4DSounds Dataset}
\label{sec:appendix_hoi_labeling}

To train the \emph{EgoTrigger} audio classifier (\Cref{sec:our_approach}), we required binary labels indicating Hand-Object Interaction (HOI) presence ($C_1$) or absence ($C_0$). We generated these labels using a subset of the Ego4DSounds dataset~\cite{chen2024action2sound}, which provides audio clips paired with time-stamped narrations from Ego4D~\cite{grauman2022ego4d}. For each sample, the Gemini 1.5 Pro model jointly analyzed the audio waveform and the corresponding narration text, interpreting both sound cues and semantic descriptions of actions to assess the likelihood of an HOI occurrence.

Based on this multimodal analysis, Gemini 1.5 Pro outputted a binary HOI label ($C_1$ or $C_0$) for every sample, enabling large-scale automated annotation. These generated labels directly facilitated the supervised training of the \emph{EgoTrigger}'s classification head, used in conjunction with the class weighting strategy detailed in \Cref{sec:our_approach}. The specific prompt provided to Gemini 1.5 Pro for this task is included below.

\section{Usage of LLMs to Perform Video QA and Judge Open-Ended QA Performance}
\label{sec:appendix_llm_qa_judge}

As detailed in~\Cref{sec:experimental_setup}, we evaluated \emph{EgoTrigger}'s downstream performance on episodic memory QA using the HME-QA and filtered QA-Ego4D~\cite{barmann2022did} datasets via an LLM-as-a-judge protocol~\cite{plizzari2025omnia, khattak2024good, maaz2023video, qian2024easy}. This involved two stages: First, the Gemini 1.5 Pro model generated an answer for each question based solely on the corresponding video segment after it had been processed by one of the four evaluated capture strategies (Full, Decimated, ET-1s, ET-Hyst.). To ensure consistency and enable fair comparisons across these strategies, this answer generation process used a fixed random seed (1337) and a low temperature setting (0.2) for deterministic outputs.

Second, the Gemini 1.5 Pro model served as the judge in a separate pass, evaluating the correctness and relevance of the answer generated in the first stage. Critically, the judge model was provided with the original question, the generated answer, and the \emph{exact same processed video segment} that the generation model used. This setup allowed the judge to assess the utility of the visual information effectively retained by each capture method based only on what was available in that specific video version. This judging phase also employed fixed seed and temperature settings (1337 and 0.2) for consistency, and each prompt to both LLMs contained only a single question-video pair to maintain evaluation focus. The specific prompt format detailing the instructions for the judge model is available below.

\onecolumn

\begin{tcolorbox}[
  colback=gray!5!white,  
  colframe=gray!75!black, 
  title=Prompt Template: Generating Memory QA Pairs from Egocentric Video,
]
\footnotesize{ 
\textbf{Task Overview:} \\
You are an AI assistant analyzing egocentric video data to generate relevant memory recall question-answer pairs.
Your task is to carefully examine the provided video (UID: \texttt{\{video\_uid\}}), its narration, and the corresponding moments annotation. Based on this information, create a list of question-answer pairs that focus on details someone might want to remember later, particularly concerning object locations, states, key interactions, task completion/confirmation, habitual actions, or notable personal experiences during the recording.

\vspace{\baselineskip} 
\textbf{Detailed Instructions:}
\begin{enumerate}
    \item \textbf{Analyze All Inputs:} Thoroughly review the video content, listen to the audio, read the narration transcript, and examine the moments annotation provided below. Synthesize information from all sources.
    \begin{verbatim}
    // Video UID: {video_uid}
    // Narration:
    {narration_pass_2}
    // Moments Annotation:
    {moments_annotation}
    \end{verbatim}

    \item \textbf{Identify Memorable Events/Details:} Focus on actions, object interactions, state changes, user experiences, and potential habits that are significant for memory recall. Avoid redundancy and pay close attention to:
    \begin{itemize}
        \item \textbf{Object Placement:} Where were objects placed or left? (e.g., "Where did I put my keys?")
        \item \textbf{Object State:} What was the final state of an object? (e.g., "Did I turn off the stove?", "Was the window closed?")
        \item \textbf{Key Interactions:} What specific action was performed with an important object? (e.g., "What did I put in the white bag?")
        \item \textbf{Task Completion/Confirmation:} Did I remember to perform a specific step or action? (e.g., "Did I remember to turn off the lights?", "Did I start the machine?")
        \item \textbf{Personal Experiences:} Did anything notable happen to me? (e.g., "Did I bump my head?")
        \item \textbf{Habitual Actions/Locations (If inferable):} Where do I normally keep things or what do I usually do in this context, if the data suggests a pattern? (e.g., "Where do I normally hang my bag?", "What do I normally add to the wash?")
        \item Focus specifically on objects, actions, and experiences mentioned or clearly visible/audible in the provided data (video, narration, annotations).
    \end{itemize}

    \item \textbf{Formulate Question-Answer Pairs:} For each identified memorable event/detail:
    \begin{itemize}
        \item \textbf{Question:} Craft a natural language question that someone might ask to recall that specific piece of information. Frame it from the perspective of the person in the video (using "I").
        \item \textbf{Answer:} Provide a concise but informative, factual answer to the question, directly supported by the video, narration, and/or moments annotation. Frame the answer using "you" as if you are informing the user (e.g., "Yes, you turned off the lights," "You left the keys on the table."). Include relevant specific details or context when available and pertinent (e.g., exact location, specific item involved).
    \end{itemize}

    \item \textbf{Determine Timestamps:} For each QA pair, identify the \texttt{start\_time\_sec} and \texttt{end\_time\_sec} from the video that correspond to the \textit{event providing the answer} to the question.
    \begin{itemize}
        \item \texttt{start\_time\_sec}: The moment the relevant action/state observation/experience \textit{begins}.
        \item \texttt{end\_time\_sec}: The moment the relevant action/state observation/experience \textit{concludes}.
        \item Format timestamps as floating-point numbers rounded to \textbf{three decimal places} (e.g., 15.000, 266.345).
    \end{itemize}
\end{enumerate}

\vspace{\baselineskip} 
\textbf{Output Format:} \\
Generate a \textbf{single JSON object} containing the video UID and a list (\texttt{'memory\_qa\_pairs'}) of the generated question-answer objects. Adhere strictly to this JSON structure:

\begin{verbatim}
{{
  "video_uid": "{video_uid}",
  "memory_qa_pairs": [
    {{
      "question": "string - Natural language question about a memorable detail (using 'I').",
      "answer": "string - Concise but informative, factual answer based on the input data, using 'you'.",
      "start_time_sec": float, // Accurate start time (seconds, rounded to 3 decimal places, e.g., 15.000)
      "end_time_sec": float  // Accurate end time (seconds, rounded to 3 decimal places, e.g., 20.000)
    }},
    // ... potentially more QA pair objects
  ]
}}
\end{verbatim}

} 
\end{tcolorbox}

\begin{tcolorbox}[
  colback=gray!5!white,
  colframe=gray!75!black,
  title=Prompt Template: Classifying HOI from Ego4DSounds Descriptions,
  label={prompt:hoi_classify} 
]
\footnotesize{ 
\textbf{Task Overview:} \\
You are an AI assistant analyzing clip descriptions from the Ego4DSounds dataset context.
Your sole task is to process a batch of clip metadata objects and determine, for each
one, if the associated \texttt{clip\_text} describes a \textbf{hand-object interaction} performed
by the camera wearer (\texttt{\#C}), according to the specific criteria below.

\vspace{\baselineskip}
\textbf{Input Data:} \\
You will be provided with a JSON array (list) containing multiple clip metadata
objects. Each object in the array represents one audio clip and contains its
\texttt{video\_uid}, \texttt{clip\_file}, \texttt{clip\_text}, and potentially other fields. The
structure for each object is expected to follow this pattern:

\begin{verbatim}
[
  {
    "video_uid": "string",        // Unique video ID
    "clip_file": "string",        // Specific audio clip filename
    "clip_text": "string",        // Text describing the action in the clip
    // ... potentially other fields like music, speech scores etc. (ignore these)
  },
  {
    "video_uid": "string",
    "clip_file": "string",
    "clip_text": "string",
     // ...
  }
  // ... potentially many clip objects total
]
\end{verbatim}

Below is the JSON array (list) for this prompt:

\texttt{\{clips\_json\}}

\vspace{\baselineskip}
\textbf{Instructions:}
\begin{enumerate}
    \item \textbf{Focus on \texttt{clip\_text}:} Carefully examine the content of the \texttt{clip\_text} field for each object. This text describes the action performed by the camera wearer (\texttt{\#C}).
    \item \textbf{Determine Interaction:} Decide if the described action primarily involves the camera wearer (\texttt{\#C}) interacting with an object using their \textbf{hands}.
    \item \textbf{Apply Specific Criteria:}
    \begin{itemize}
        \item \textbf{Consider it hand-object interaction (classify as \texttt{1}) if:} The \texttt{clip\_text} clearly describes the person manipulating, holding, picking up, placing, using, touching, opening, or closing an object with their hands. Examples: "\#C picks a bag", "\#C opens the drawer", "\#C uses a spoon", "\#C types on keyboard", "\#C washes hands", "\#C writes on paper". Include actions that strongly imply hand use for the primary interaction.
        \item \textbf{Do NOT consider it hand-object interaction (classify as \texttt{0}) if:} The \texttt{clip\_text} describes whole-body motion without specific hand-object focus (e.g., "\#C walks out", "\#C stands up", "\#C sits down"), passive observation, environmental descriptions, or actions explicitly performed \textit{without} hands (e.g., "\#C opens the cabinet door with leg", "\#C kicks the ball"). If the primary action doesn't involve hands on an object, classify as \texttt{0}.
    \end{itemize}
    \item \textbf{Output Classification:} Based on the above, determine a single binary integer value (\texttt{1} if it describes hand-object interaction, \texttt{0} otherwise) for each clip object.
\end{enumerate}

\vspace{\baselineskip}
\textbf{Output Format:} \\
Generate \textbf{ONLY} a single, valid JSON array (list) containing results for all
clip objects provided in the input batch. Each object within the array must
contain the \texttt{video\_uid}, \texttt{clip\_file}, \texttt{clip\_text}, and the corresponding binary
classification (\texttt{is\_hand\_object\_interaction}). Adhere strictly to this format,
with no other text before or after the JSON array block:

\begin{verbatim}
[
  {
    "video_uid": "string_id_of_first_clip",
    "clip_file": "filename_of_first_clip",
    "clip_text": "text_description_of_first_clip",
    "is_hand_object_interaction": 1_or_0_value_for_first_clip
  },
  {
    "video_uid": "string_id_of_second_clip",
    "clip_file": "filename_of_second_clip",
    "clip_text": "text_description_of_second_clip",
    "is_hand_object_interaction": 1_or_0_value_for_second_clip
  }
  // ... include one object for each clip in the input batch
]
\end{verbatim}

} 
\end{tcolorbox}

\begin{tcolorbox}[
  colback=gray!5!white,
  colframe=gray!75!black,
  title=Prompt Template: Answering Question based on Video Clip,
  label={prompt:video_qa_gen} 
]
\footnotesize{ 
\textbf{Task:} Analyze the provided video clip and answer the single question below based \textit{only} on the visual and auditory information within the clip.

\vspace{\baselineskip}
\textbf{Input Video:} (Provided separately) \\
\textbf{Video UID:} \texttt{\{video\_uid\}}

\vspace{\baselineskip}
\textbf{Question:} \\
\texttt{\{current\_question\}}

\vspace{\baselineskip}
\textbf{Instructions:}
\begin{enumerate}
    \item Carefully watch and listen to the entire video clip.
    \item Provide a concise and factual answer to the specific question asked above, based \textit{solely} on the content of the video.
    \item Do not infer information not present in the video. If the answer cannot be determined from the video, state that clearly (e.g., "Cannot be determined from the video.").
    \item Format your response as a single JSON object containing the original "question" and the corresponding "answer" you generated.
\end{enumerate}

\vspace{\baselineskip}
\textbf{Output Format (Strict JSON):}
\begin{verbatim}
{{
  "question": "{current_question_escaped}",
  "answer": "string - Your answer to the specific question based on the video."
}}
\end{verbatim}

} 
\end{tcolorbox}

\begin{tcolorbox}[
  colback=gray!5!white,
  colframe=gray!75!black,
  title=Prompt Template: Judging Video QA Answer Correctness,
  label={prompt:video_qa_judge} 
]
\footnotesize{ 
\textbf{ROLE:} You are an objective AI evaluator. Your task is to assess the correctness and quality of a predicted answer compared to a ground-truth answer for a given video-based question.

\vspace{\baselineskip}
\textbf{INSTRUCTIONS:}
\begin{enumerate}
    \item Carefully read the Question, Ground Truth Answer, and the Predicted Answer.
    \item Compare the Predicted Answer directly against the Ground Truth Answer.
    \item Focus on factual correctness and semantic equivalence. Minor phrasing differences are acceptable if the core meaning is identical.
    \item \textbf{Score} the Predicted Answer on an integer scale from 0 to 5:
        \begin{itemize}
            \item \textbf{5: Perfect Match.} The predicted answer is factually correct and semantically identical or a trivial rephrasing of the ground truth.
            \item \textbf{4: Mostly Correct.} The predicted answer is factually correct but might miss a minor detail or have slightly different phrasing that is still accurate.
            \item \textbf{3: Partially Correct.} The predicted answer contains significant factual overlap with the ground truth but is missing key elements or includes some minor inaccuracies.
            \item \textbf{2: Minimally Correct / Related.} The predicted answer touches upon the topic but is largely incorrect or misses the main point of the ground truth.
            \item \textbf{1: Incorrect but Relevant.} The predicted answer is factually incorrect but addresses the question's topic.
            \item \textbf{0: Completely Incorrect / Irrelevant.} The predicted answer is factually wrong and/or does not address the question meaningfully.
        \end{itemize}
    \item \textbf{IMPORTANT:} Consider uncertain predictions (e.g., "I cannot determine", "It is impossible to answer") as \textbf{incorrect (Score 0)}, \textit{unless} the Ground Truth Answer \textit{also} explicitly states uncertainty or inability to answer.
    \item Provide a brief \textbf{Reason} explaining your score, highlighting agreements or discrepancies between the prediction and the ground truth.
    \item Determine the overall \textbf{Judgement} as 'correct' (typically scores 3, 4, or 5) or 'incorrect' (typically score 0-2). Use 'correct' only if the prediction fully captures the essential information of the ground truth.
\end{enumerate}

\vspace{\baselineskip}
\textbf{INPUT:}
\begin{itemize}
    \item \textbf{Question:} \texttt{\{question\}}
    \item \textbf{Ground Truth Answer:} \texttt{\{ground\_truth\_answer\}}
    \item \textbf{Predicted Answer:} \texttt{\{predicted\_answer\}}
\end{itemize}

\vspace{\baselineskip}
\textbf{REQUIRED OUTPUT FORMAT (Strict JSON):} \\
Output \textit{only} a single JSON object in the following format. Do not include any text before or after the JSON block.

\begin{verbatim}
{{
  "video_id": "{video_id}",
  "question": "{question_escaped}",
  "ground_truth_answer": "{ground_truth_answer_escaped}",
  "predicted_answer": "{predicted_answer_escaped}",
  "judgement": "string - 'correct' or 'incorrect'",
  "score": integer,
  "reason": "string - Brief explanation for the score."
}}
\end{verbatim}

} 
\end{tcolorbox}

\twocolumn